\documentclass[11pt]{article}
\usepackage{acl}
\usepackage{times}
\usepackage{latexsym}
\usepackage[T1]{fontenc}
\usepackage[utf8]{inputenc}
\usepackage{microtype}
\usepackage{graphicx}
\usepackage{amsmath,amssymb}
\usepackage{booktabs}
\usepackage{multirow}
\usepackage{float}
\usepackage{mdframed}
\mdfdefinestyle{codestyle}{backgroundcolor=black!4,linecolor=black!25,leftline=true,rightline=false,topline=false,bottomline=false,linewidth=1.2pt,innerleftmargin=5pt,innerrightmargin=3pt,innertopmargin=3pt,innerbottommargin=3pt,skipabove=6pt,skipbelow=6pt}
\mdfdefinestyle{quotestyle}{backgroundcolor=blue!3,linecolor=blue!35,leftline=true,rightline=false,topline=false,bottomline=false,linewidth=1.2pt,innerleftmargin=5pt,innerrightmargin=3pt,innertopmargin=3pt,innerbottommargin=3pt,skipabove=5pt,skipbelow=5pt}

\title{Pull: Lazy Materialization of Working Memory for Stateful LLM Conversations}

\author{Jiangang Chen\thanks{Chengdu Beiluoshimen Technology Co., Ltd.}}

\begin{document}
\maketitle

\begin{abstract}
As LLM conversations grow to hundreds of turns, full-context injection incurs $O(N^2)$ cumulative token costs, while lossy summarization or hard truncation irreversibly discards historical state. We propose \textbf{Pull}, a session router that maintains an addressable metadata directory via a local, deterministic Purifier (zero LLM calls, millisecond-level latency). At query time, the LLM lazily materializes only the turns it needs; unmaterialized turns remain accessible but collapsed. Unlike irreversible compression, Pull's materialization is reversible---subsequent queries can expand any collapsed turn. On LoCoEval (128 conversations, 12{,}780 turns), Pull reduces per-query context tokens (Phase~2) by 75.1\% on single-hop tasks with equivalent quality ($\Delta = -0.002$, n.s.) and by 72.0\% on multi-hop tasks with no quality loss ($\Delta = +0.017$). A controlled routing benchmark (7{,}831 queries $\times$ 10 methods) shows that entity lifecycle tracking is empirically a prerequisite for distance-independent routing. On BEAM~1M (14 conversations, 263 questions), Pull improves F1 by +55.2\% over a truncation baseline.\footnote{Code and reproduction scripts are available at \url{https://github.com/wulun811/kongmen-pull}; evaluation benchmarks (LoCoEval, BEAM) are external and linked from the repository README.}
\end{abstract}

\section{Introduction}\label{sec:intro}

When LLM conversations grow to hundreds of turns, developers face a cost--fidelity tradeoff. Full-context injection---the default behavior of all mainstream LLM APIs---causes per-request context length to grow linearly ($O(N)$). Over a complete session lifecycle, cumulative retransmitted tokens grow quadratically ($O(N^2)$).

An exploratory analysis of 287 coding-assistant sessions (59{,}520 messages, 5.56 billion billed tokens) reveals a ${\sim}$20$\times$ token amplification (5.56B billed / 280M original $\approx$ 19.9$\times$): only 280 million tokens are original content, yet APIs bill 5.56 billion. Entity overlap between queries and history has a median of only 20.7\%, meaning the model is forced to search through largely irrelevant history.

At the other extreme, lossy summarization \citep{jiang2023llmlingua,li2023selective} or hard truncation irreversibly discards historical details. Once a turn is summarized away, its precise state---variable definitions, branch decisions, deprecated implementations---is permanently lost.

Our experiments provide strong evidence: by lazily materializing only query-relevant turns, the input signal-to-noise ratio improves substantially, yielding a +55.2\% Overall F1 improvement on million-token conversations (BEAM~1M, $\Delta = +0.165$ vs.\ truncation baseline). This suggests that the ``Lost in Conversation'' phenomenon \citep{laban2025lost} may stem not from Transformer attention itself, but from the input signal-to-noise ratio.

Engineering approaches using RAG \citep{lewis2020rag} or long-term memory systems \citep{packer2023memgpt,chhikara2025mem0} excel at cross-session knowledge archival but have architectural limitations for \emph{intra-session} state management. RAG returns text fragments divorced from temporal causality; memory systems may lose branch topology and entity lifecycles. In short, the first layer (full context) manages ``the current query'' and the third layer (long-term memory) manages ``what happened in past sessions''; but the second layer---a live session state router---lacks a clear system abstraction.

We propose \textbf{Pull}, a session router that fills this gap. Pull redefines session management as \emph{lazy materialization}: rather than pushing compressed summaries or retrieving dead fragments, it maintains a live metadata directory. The LLM is given this directory---containing entity lifecycles, branch topology, and causal dependencies---and actively \emph{pulls} the specific turns it needs into working memory. Unmaterialized turns remain accessible but collapsed; subsequent queries can lazily expand them.

Pull's online metadata construction is performed by a local, deterministic component: the \textbf{Purifier} computes structured metadata per turn (entity extraction, lifecycle state machine, branch ID, fidelity score) using lightweight ONNX embeddings and rule engines---zero LLM calls. It maintains a \textbf{Directory Layer (M0)} (per-turn extractive summaries, ${\sim}$100 chars each) and a \textbf{Trace Layer (M10)} (entity-level dependency graph with lifecycle labels and branch topology).

At query time, a \textbf{Selector} presents the metadata tree to the LLM; the LLM then decides, in a single call, which turns to materialize, triggering cascade expansion to preserve causal chains. The \textbf{Answerer} receives materialized turn originals plus the global M0 directory.

Our contributions are:
\begin{enumerate}
    \item We identify and formalize the missing layer in LLM session infrastructure---the session router (second layer)---and propose lazy materialization as its governing paradigm.
    \item We design Pull, a deterministic metadata-driven session router that maintains branch topology, entity lifecycles, and causal chains for 300+ turn conversations using only local lightweight computation.
    \item On LoCoEval (128 conversations, 12{,}780 turns), Pull reduces Phase~2 tokens by 75.1\% (single-hop) and 72.0\% (multi-hop) with equivalent or improved quality. On BEAM~1M, Pull improves F1 by +55.2\% over truncation.
    \item A controlled routing benchmark (7{,}831 queries $\times$ 10 methods) with five-level ablation shows entity lifecycle tracking is empirically a prerequisite for distance-independent routing; LLM Selector validation confirms a 3.9$\times$ improvement of M10 over M0.
\end{enumerate}

\section{Related Work}

\paragraph{Hard truncation and sliding windows.} Last-K, vLLM PagedAttention \citep{kwon2023pagedattention}, and industrial defaults operate on the first layer by irreversibly discarding history. Once truncated, precise state is permanently lost and causal chains break.

\paragraph{Lossy summarization and compression.} LLMLingua \citep{jiang2023llmlingua}, ICAE \citep{ge2024icae}, LangChain SummaryMemory \citep{langchain2023}, and Selective Context \citep{li2023selective} shorten prompts through irreversible merging or token dropping. These methods optimize input format but destroy the information substrate.

\paragraph{Reversible context orchestration.} ACE \citep{liao2026ace} likewise targets reversibility: it retains raw agent-trajectory steps alongside LLM-generated abstractions and re-decides at every decision step whether each historical step is presented as raw, abstract, or dropped. R$^3$Mem \citep{wang2025r3mem} reconstructs compressed context by invoking the model backward. Pull differs from both in two ways: (i) its metadata layer is produced by a deterministic Purifier with zero LLM calls---entity lifecycles, fidelity grades, and branch topology---whereas ACE's abstractions are LLM summaries without entity-level structure; (ii) Pull targets intra-session conversational working memory with turn-level materialization, whereas ACE orchestrates agent trajectory steps.

\paragraph{Retrieval-augmented and cross-session archival.} Mem0 \citep{chhikara2025mem0}, Zep \citep{rasmussen2025zep}, vector RAG \citep{lewis2020rag}, GraphRAG \citep{edge2024graphrag}, and HippoRAG \citep{gutierrez2024hipporag} index conversation history for recall, primarily addressing cross-session archival. While retrieval augmentation can reduce hallucination \citep{shuster2021rag}, retrieved fragments lack temporal causality. We use BM25 \citep{robertson2009bm25} and Dense RAG (ONNX MiniLM \citep{reimers2019sentencebert}) as minimal instances of this paradigm.

\paragraph{LLM-driven memory paging.} MemGPT/Letta \citep{packer2023memgpt,letta2024} simulates the second layer via virtual memory management but delegates paging decisions to the LLM itself. Pull differs by focusing on turn-level working memory routing with deterministic metadata, rather than agent state and long-term memory runtime.

\paragraph{Multi-turn degradation.} \citet{laban2025lost} confirm that all mainstream LLMs degrade by 39\% on average in multi-turn conversations. LongMemEval \citep{wu2025longmemeval} finds 30\% accuracy drops in sustained interaction. These works quantify degradation but do not propose system-level solutions. Pull addresses this architecturally by maintaining the signal-to-noise ratio above a threshold via lazy materialization.

\paragraph{Companion work.} The Purifier builds on our companion work KongMen \citep{chen2026kongmen}, which introduces training-free entity-redundancy gating for real-time dialogue purification. Pull extends this line from turn-level purification to session-level lazy materialization with entity lifecycle tracking.

\section{Method}
\label{sec:method}

\subsection{Problem Formulation}

Let $C = [t_1, t_2, \ldots, t_N]$ be a conversation of $N$ turns. Given query $q$, the goal is to select a subset $S \subseteq \{1,\ldots,N\}$ minimizing token count while bounding information loss:
\begin{equation}
\begin{aligned}
\arg\min_S \quad & \text{TokenCount}(\text{prompt}(q, S)) \\
\text{s.t.} \quad & \mathcal{L}(\text{answer}(q, S), \text{answer}(q, C)) \leq \varepsilon
\end{aligned}
\end{equation}
Full history uses $S = \{1,\ldots,N\}$; Pull finds an $S \ll N$. In practice, $\text{answer}(q,C)$ is intractable; Pull uses a computable proxy: $S = f_\theta(q, M_0, M_{10})$, where $f_\theta$ is the Selector.

\subsection{Architecture Overview}

\begin{figure*}[t]
\centering
\includegraphics[width=\textwidth]{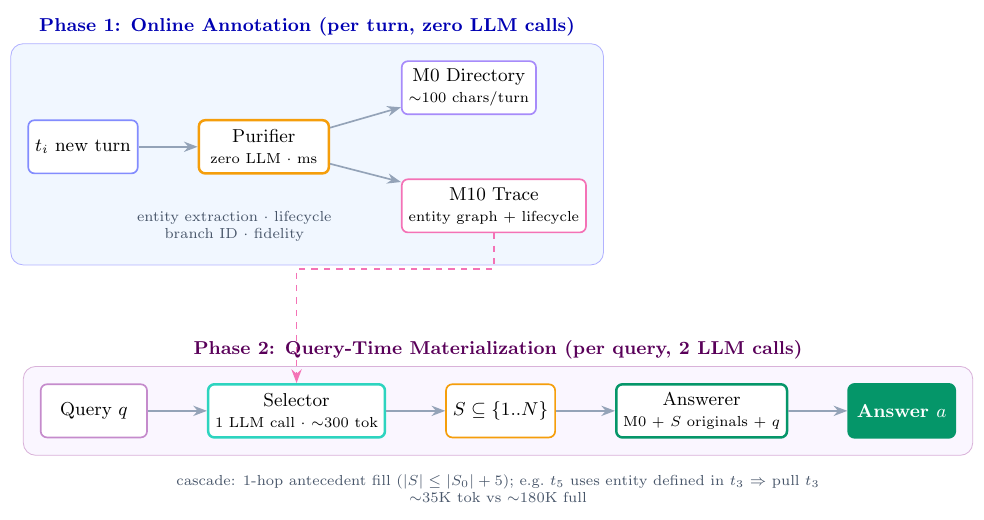}
\caption{Pull architecture. Phase~1 (online): the Purifier annotates each turn with structured metadata (M0 directory + M10 trace layer) using zero LLM calls. Phase~2 (query): the Selector reads M10 to pick relevant turns $S$; cascade expansion fills missing causal dependencies; the Answerer receives M0 + $S$ originals.}
\label{fig:architecture}
\end{figure*}

Pull operates in two orthogonal phases: \textbf{online annotation} (per turn, zero LLM calls) and \textbf{query-time materialization} (per query, 2 LLM calls). The key insight is that these two phases are fully decoupled in both time and computation (Figure~\ref{fig:architecture}).

\begin{figure*}[t]
\centering
\includegraphics[width=\textwidth]{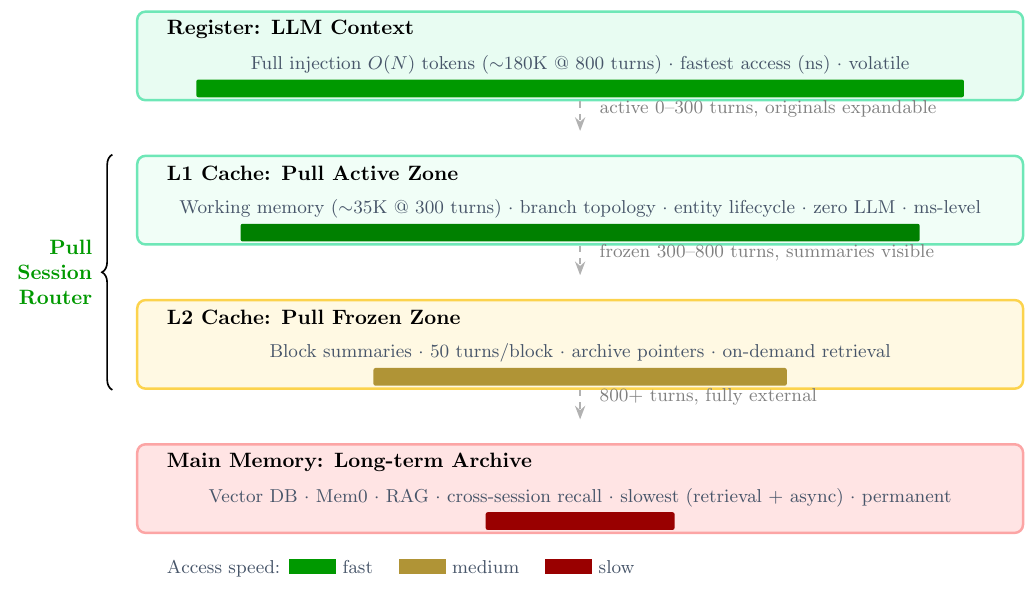}
\caption{Three-tier memory hierarchy. Pull occupies the second layer (L1+L2 cache): deterministic local metadata for active turns, block summaries for frozen turns. Unlike irreversible compression, collapsed turns remain expandable.}
\label{fig:tiered}
\end{figure*}

\paragraph{Three-tier storage.} Active zone (0--300 turns): originals + metadata. Frozen zone (300--800 turns): block summaries + metadata + archive pointers. External archive: third-party storage. This mirrors the classical OS memory hierarchy (Figure~\ref{fig:tiered}).

\paragraph{M0/M10 division of labor.} M0 (Directory Layer) is a global summary directory (${\sim}$100 chars/turn); the Answerer always reads the full M0 for ``positional awareness.'' M10 (Trace Layer) is the entity dependency graph + lifecycle labels + branch topology; the Selector reads M10 for routing decisions without reading originals.

\subsection{Online Annotation: Purifier}

For each turn $t_i$, the Purifier computes a metadata tuple $m_i$ comprising six fields: extractive summary, entity set, information density, fidelity score, lifecycle state, and branch ID. Entity extraction compiles 16 regex patterns grouped into 13 type families (function/class definitions, code identifiers such as snake\_case/camelCase/backtick-quoted names, URLs, and temporal/numeric/named entities); e.g.\ \texttt{\textbackslash b(?:def|class)\textbackslash s+(\textbackslash w+)} captures definition names and \texttt{[a-z]+\_[a-z]+} captures snake\_case identifiers (full list in Appendix~\ref{sec:purifier-detail}). Lifecycle tracking is a 4-state machine over $\Sigma=\{\textsc{def},\textsc{ref},\textsc{mod},\textsc{dep}\}$ whose transitions are deterministic functions of entity position and code-signature hashes (Table~\ref{tab:lifecycle}); \textsc{dep} and \textsc{mod} are mutually exclusive (an entity absent from the current turn cannot be modified in it). Fidelity is a rule-based scorer combining nine pragmatic and lexical signals (Appendix~\ref{sec:purifier-detail}). The architecture treats entity extraction as a pluggable domain adapter: the 13 templates target code conversations, and other domains substitute their own extractor without changing the Selector/Answerer pipeline (Limitations).

\begin{table}[t]
\centering
\small
\setlength{\tabcolsep}{4pt}
\begin{tabular}{@{}p{2.5cm}p{4.9cm}@{}}
\toprule
\textbf{Transition} & \textbf{Deterministic trigger} \\
\midrule
$\to$\textsc{def} & entity's first-occurrence turn \\
\textsc{def}$\to$\textsc{ref} & re-occurrence after the first turn \\
$\to$\textsc{mod} & code-signature hash (def/class lines) changes \\
$\to$\textsc{dep} & was code-defined, absent now, replaced via a bridge entity \\
\bottomrule
\end{tabular}
\caption{Lifecycle state machine. All triggers are rule-based (entity position, signature hash, co-occurrence); no LLM is involved.}
\label{tab:lifecycle}
\end{table}

All computation is local (MiniLM ONNX ${\sim}$46MB + regex + rules). Measured on a single machine (Intel i5-9400F, 6 cores, 2.90\,GHz, 23.4\,GB RAM, Python 3.12.3), processing an 800-turn conversation incrementally (one turn at a time): P50 = 3.22 ms/turn, P99 = 32.5 ms; session-state memory grows ${\sim}$linearly to ${\sim}$30\,MB at 800 turns.

\subsection{Eviction and Archival}

When active turns exceed 300, lowest-priority turns are evicted based on weighted signals (fidelity, reference count, content length, recency). Evicted turns are frozen into block summaries stored in M0's archive zone. Query-time protection temporarily raises retention priority for turns relevant to the current query.

\subsection{Query-Time Materialization: Selector}

The Selector takes M10 metadata + query $q$ and outputs a turn set $S$. Concretely, it receives M10 as structured plain text---one line per turn carrying entity, lifecycle, density, and branch fields (Appendix~\ref{sec:selector-prompt})---and emits, in a single LLM call, a one-sentence reason plus a comma-separated turn list; the cascade expansion below is deterministic post-processing applied \emph{after} the LLM output, not an LLM decision. The pipeline:
\begin{enumerate}
    \item Parse entities from $q$.
    \item Match relevant turns in M10.
    \item Fidelity check.
    \item \textbf{Cascade expansion}: for every entity referenced in the LLM's initial set $S_0$, the turn of its earliest definition is added if not already selected. Expansion is one hop (no transitive closure) and bounded by a fixed budget of 5 extra turns ($|S| \leq |S_0| + 5$).
    \item Validate output (range check, dedup).
    \item Fallback to entity matching + M0 similarity on parse failure.
\end{enumerate}

\subsection{Query-Time Materialization: Answerer}

The Answerer receives the full M0 directory + original text of turns in $S$ + query $q$, and generates the answer. M0 summaries are extractive (MiniLM-based + rule extraction) and involve no LLM generation, thus ensuring factual consistency.

\section{Experimental Setup}

\paragraph{LoCoEval} \citep{liu2026locoeval} is the first benchmark for repository-level long-horizon conversational context management: 128 samples (64 single-hop + 64 multi-hop) from 71 real open-source repositories, 60--138 turns each (mean 100), Phase~2 context 35K--224K tokens. Three tasks: Information Extraction (IE), Topic Awareness (TA), Function Completion (FC). We focus on IE+TA (256 instances); FC requires a pytest execution environment and is excluded.

\paragraph{BEAM 1M} \citep{tavakoli2025beam} is a public long-conversation memory benchmark: 35 conversations, 700 questions. We select 14 conversations to control topic diversity rather than maximize count (263 questions). Conversations are 700K--1.8M tokens, exceeding LLM context limits.

\paragraph{Controlled routing benchmark.} We stitch 60 SWE-bench traces (code domain) into 12 sessions (140--1407 steps) and derive 7{,}831 probe questions with deterministic ground truth.\footnote{Ten zero-LLM routing methods, each returning top-10 candidates: Oracle (upper bound), M10-Entity, RAG+Entity, RAG+Timestamp, RAG-Hybrid (BM25+dense, RRF), RAG-BM25, RAG-Dense, M0-BM25, M0-Keyword, and Last-K ($K{=}30$).} Three layers: (A) general retrieval (53\%), (B) state-change tracking (22\%), (C) adversarial (25\%).

\paragraph{Implementation.} Generation and judge model: GLM-5, temperature = 0.0. Pull Selector uses M10 + selection prompt; Answerer uses M0 + expanded turns + answerer prompt.

\paragraph{Metrics.} IE/TA F1 via LLM-as-a-Judge; FC pass@1 via pytest. Token savings: $1 - \tfrac{\sum T_{\text{Pull}}}{\sum T_{\text{Full}}}$ (Phase~2 only).

\section{Experiments}\label{sec:experiments}

\subsection{Single-Hop Main Results}

Table~\ref{tab:singlehop} shows Pull vs.\ baselines on LoCoEval single-hop.

\begin{table}[t]
\centering
\small
\begin{tabular}{@{}lcccc@{}}
\toprule
\textbf{Method} & \textbf{F1} & $\Delta$ & \textbf{Tokens} & \textbf{Save} \\
\midrule
Vanilla & 0.6694 & --- & 95{,}406 & --- \\
\textbf{Pull} & \textbf{0.6676} & $\mathbf{-0.002}$ & \textbf{23{,}719} & \textbf{75.1\%} \\
RAG (BM25) & 0.3562 & $-$0.313 & 19{,}557 & 79.5\% \\
RAG (Dense) & 0.4001 & $-$0.269 & 16{,}370 & 82.8\% \\
Random-K & 0.5572 & $-$0.112 & 20{,}721 & 78.3\% \\
Last-K & 0.3044 & $-$0.365 & 22{,}276 & 76.7\% \\
\bottomrule
\end{tabular}
\caption{LoCoEval single-hop results (64 conversations, batched Answerer). Pull uses batched Answerer (one Selector + one Answerer for all tasks); baselines use per-task calls.}
\label{tab:singlehop}
\end{table}

\textbf{Equivalence.} $\Delta = -0.0018$ is far below judge noise (mean per-sample std.\ dev.\ = 0.018). TOST (Two One-Sided Tests) passes at the $\pm$0.05 boundary ($p < 0.01$), confirming quality equivalence with 75.1\% token savings. To rule out same-model evaluation bias, an independent judge from a different model family (Qwen3.7plus) confirms directional consistency ($\Delta = +0.0148$, Pearson $r > 0.85$ on sample ordering; Table~\ref{tab:judge}). Although 22/64 individual samples flip direction, the near-identical sample \emph{ranking} places the Pull--Vanilla gap within the judge's noise floor rather than indicating a systematic quality difference.

\begin{figure*}[t]
\centering
\includegraphics[width=\textwidth]{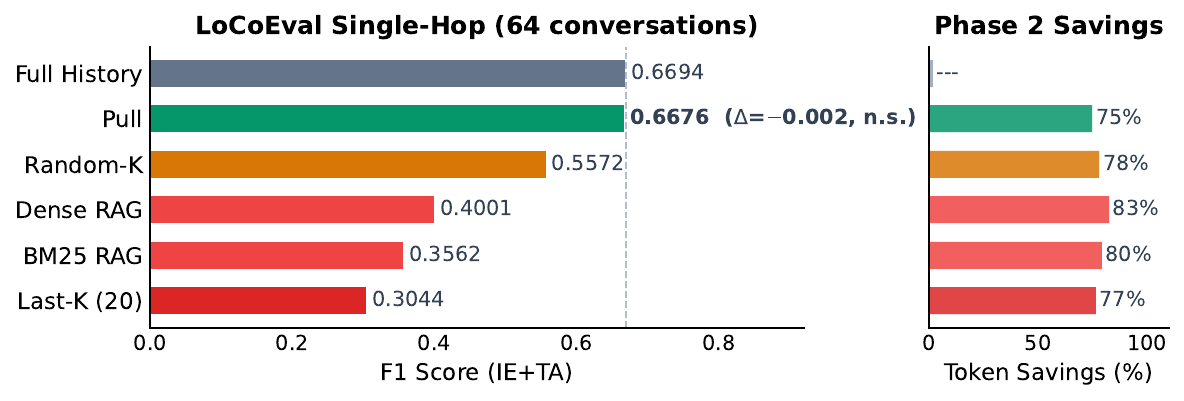}
\caption{LoCoEval single-hop F1 and Phase~2 token savings. Pull matches full history (dashed line) while saving 75.1\% tokens. RAG and truncation baselines suffer catastrophic quality loss.}
\label{fig:mainresults}
\end{figure*}

\subsection{Ablation}

\begin{table}[t]
\centering
\small
\begin{tabular}{@{}lcccc@{}}
\toprule
\textbf{Variant} & \textbf{F1} & $\Delta$ vs Van. & \textbf{Tokens} & \textbf{Save} \\
\midrule
Vanilla & 0.6694 & --- & 95{,}406 & --- \\
Full Pull & 0.6676 & $-$0.002 & 23{,}719 & 75.1\% \\
w/o Cascade & 0.6728 & +0.003 & 22{,}736 & 76.2\% \\
w/o M0 & 0.6103 & $-$0.059 & 22{,}578 & 76.3\% \\
\bottomrule
\end{tabular}
\caption{Ablation (64 conversations, single-hop, batched Answerer).}
\label{tab:ablation}
\end{table}

\textbf{w/o M0 drops significantly} ($-$0.059, well beyond judge noise $\sigma=0.018$), confirming M0 is critical for the Answerer's positional awareness in batched mode. Cascade has only a marginal, non-significant effect in single-hop ($\Delta=+0.003$, within $\sigma=0.018$; short chains are already covered by initial matching).

\subsection{Multi-Hop Results}

\begin{table}[t]
\centering
\small
\begin{tabular}{@{}lcccc@{}}
\toprule
\textbf{Method} & \textbf{F1} & \textbf{Tokens} & \textbf{Save} & $\Delta$ vs Pull \\
\midrule
Vanilla$^\dagger$ & 0.6326 & 86{,}118 & --- & --- \\
Full Pull & \textbf{0.6500} & 24{,}115 & \textbf{72.0\%} & --- \\
w/o Cascade & 0.6372 & 24{,}092 & 72.0\% & $-$0.013 \\
w/o M0 & 0.6331 & 22{,}945 & 73.4\% & $-$0.017 \\
\bottomrule
\end{tabular}
\caption{Multi-hop results (64 conversations). $^\dagger$Vanilla F1 from 32 overlapping samples.}
\label{tab:multihop}
\end{table}

Pull achieves 72.0\% token savings (0/64 negative) with quality on par with full injection. The cascade contribution is larger in multi-hop ($\Delta = +0.013$) than single-hop ($\Delta = -0.005$), consistent with increasing cross-segment dependencies. Independent judge (Qwen3.7plus) confirms direction ($\Delta = +0.011$).

\subsection{BEAM 1M Long-Conversation Validation}

BEAM~1M conversations (700K--1.8M tokens) exceed GLM-5's context window. The truncation baseline uses Last-K hard truncation (${\sim}$163K tokens). Pull uses a simplified Purifier (M0 directory only, without the M10 lifecycle layer). This is deliberate: BEAM isolates the \emph{lazy-materialization paradigm}---directory-guided on-demand expansion---from the lifecycle-routing contribution validated independently in \S\ref{sec:routing}. The two experiments answer orthogonal questions: BEAM tests whether lazy materialization beats truncation at million-token scale; \S\ref{sec:routing} tests which metadata signals make routing distance-independent.

\begin{table}[t]
\centering
\small
\begin{tabular}{@{}lccc@{}}
\toprule
\textbf{Metric} & \textbf{Trunc.} & \textbf{Pull} & $\Delta$ \\
\midrule
Overall F1 & 0.299 & \textbf{0.464} & \textbf{+55.2\%} \\
Win/Tie/Loss & --- & 122/118/23 & --- \\
Token mean & 163K & \textbf{64.5K} & 60.4\% \\
\bottomrule
\end{tabular}
\caption{BEAM 1M results (14 conversations, 263 questions).}
\label{tab:beam}
\end{table}

\begin{table}[t]
\centering
\small
\resizebox{\columnwidth}{!}{%
\begin{tabular}{@{}lccccc@{}}
\toprule
\textbf{Variant} & \textbf{Trunc.} & \textbf{Pull} & $\Delta$ & \textbf{Tok.} & \textbf{Save} \\
\midrule
Full Pull & 0.299 & \textbf{0.464} & +0.165 & 64{,}517 & 60.4\% \\
w/o Cascade & 0.298 & 0.447 & +0.149 & 39{,}404 & 75.8\% \\
w/o M0 & 0.293 & 0.428 & +0.135 & 54{,}154 & 66.8\% \\
\bottomrule
\end{tabular}}
\caption{BEAM 1M ablation (263 questions).}
\label{tab:beam_ablation}
\end{table}

At comparable token budget (${\sim}$65K), Pull (F1=0.464) outperforms the best retrieval baseline (Hybrid, 63K tokens) by $+0.336$ (Table~\ref{tab:beam_retrieval}). Even when baselines receive 2.5$\times$ more tokens (163K), they reach only F1=0.137 ($-$0.327).

\subsection{Controlled Routing: Structured Metadata Ablation}
\label{sec:routing}

\begin{table}[t]
\centering
\small
\begin{tabular}{@{}lcccl@{}}
\toprule
\textbf{Method} & \textbf{R@10} & \textbf{MRR} & \textbf{Incr.} & \textbf{Index} \\
\midrule
RAG-Dense & 46.4\% & 0.239 & --- & MiniLM dense \\
RAG-BM25 & 55.3\% & 0.335 & +8.9 & Raw text \\
+Timestamp & 71.3\% & 0.495 & +16.0 & +step prefix \\
+Entity & 74.1\% & 0.514 & +2.8 & +struct.\ fields \\
\textbf{M10} & \textbf{96.0\%} & \textbf{0.825} & \textbf{+21.9} & +lifecycle \\
\bottomrule
\end{tabular}
\caption{Five-level structured metadata ablation (7{,}831 queries, Recall@10).}
\label{tab:routing_ablation}
\end{table}

Table~\ref{tab:routing_ablation} reveals three levels of contribution:
(1)~BM25 exact matching outperforms dense retrieval (55.3\% vs 46.4\%);
(2)~timestamps and structured entity fields raise Recall from 55.3\% to 74.1\%;
(3)~M10's entity lifecycle tracking (def/mod/dep) provides an additional +21.9\%, pushing Recall to 96.0\%.
\textbf{Lifecycle tracking is empirically a prerequisite for distance-independent routing in this benchmark}---without it, Recall plateaus at 74\%.

M10 shows no distance decay: Recall stays at 78--98\% from $\leq$30 steps to $>$200 steps, while Last-K collapses from 75.6\% to 0.8\% and all RAG variants decay with distance. LLM Selector validation (198 sampled queries) confirms M10 achieves 69.9\% vs M0's 17.9\% (3.9$\times$ improvement).

\subsection{Comparison with LLMLingua2}

\begin{table}[t]
\centering
\small
\begin{tabular}{@{}lcccc@{}}
\toprule
\textbf{Method} & \textbf{F1} & \textbf{Tokens} & \textbf{Save} & \textbf{Recov.} \\
\midrule
Vanilla$^\dagger$ & 0.6326 & 86{,}118 & --- & \checkmark \\
\textbf{Pull} & \textbf{0.6500} & \textbf{24{,}115} & \textbf{72.0\%} & \checkmark \\
LLMLingua2 (r50) & 0.6118 & 30{,}516 & 64.6\% & $\times$ \\
LLMLingua2 (r28) & 0.5079 & 15{,}310 & 82.2\% & $\times$ \\
\bottomrule
\end{tabular}
\caption{Pull vs LLMLingua2 \citep{pan2024llmlingua2} (multi-hop, 64 conversations). $^\dagger$32-sample subset.}
\label{tab:llmlingua}
\end{table}

Pull outperforms LLMLingua2 at both compression ratios: $\Delta = +0.038$ (vs r50) and $+0.142$ (vs r28). The key architectural difference: LLMLingua2's compression is \textbf{irreversible}, while Pull's materialization is \textbf{reversible}---unmaterialized turns remain accessible for subsequent queries.

A reversibility probe on 16 multi-hop samples confirms this: Pull-Adaptive (per-task independent routing) reaches F1=0.6785, whereas LLMLingua2-Fixed (one compression shared across tasks) reaches only F1=0.5272 ($\Delta = +0.151$), demonstrating that adaptive routing is the core source of Pull's quality advantage. The comparison is multi-hop only: single-hop queries localize to a single turn, and even retrieval baselines collapse there (RAG F1 $\leq$ 0.40, Table~\ref{tab:singlehop}), so compression robustness is not a differentiating factor on that split (Limitation~3).

\section{Discussion}

\subsection{Cost Analysis: Model Tiering}

Pull's two-phase design enables \textbf{model tiering}: the Selector needs only ${\sim}$8K tokens of structured metadata, allowing a cheap model; the Answerer handles originals with a flagship model. Full injection has no such tiering.

\begin{table}[t]
\centering
\small
\resizebox{\columnwidth}{!}{%
\begin{tabular}{@{}lccc@{}}
\toprule
\textbf{Configuration} & \textbf{Selector cost} & \textbf{Total cost/query} & \textbf{vs Full} \\
\midrule
Full (Fable 5) & --- & \$0.91 & --- \\
Pull same-model & \$0.095 & \$0.39 & $-$58\% \\
Pull tiered (DS+Fable) & \$0.001 & \$0.29 & $-$68\% \\
Full (Kimi K3) & --- & \$1.37 & --- \\
Pull tiered (DS+K3) & \$0.001 & \$0.44 & $-$68\% \\
\bottomrule
\end{tabular}}
\caption{Per-query cost (100-turn conversation, multi-hop, 2026-07 pricing). Claude Fable 5: \$10/\$50 per MTok; Kimi K3: \$15/\$75; DeepSeek V4 Flash: \$0.14/\$0.28. Token counts from Table~\ref{tab:multihop}.}
\label{tab:cost}
\end{table}

In multi-hop (72\% savings), Pull reduces per-query cost from \$0.91 to \$0.29 ($-$68\%). Model tiering transforms Selector cost from $O(N \times P_{\text{expensive}})$ to $O(N \times P_{\text{cheap}})$, where $P_{\text{cheap}}/P_{\text{expensive}} \approx 1/70$.

\subsection{Production Deployment}

Pull is integrated into a production coding assistant, running across 13 SWE-bench Pro instances (350+ tool rounds, zero crashes).

The M0 directory effectively prevents redundant LLM searches: in one session, the LLM skipped directories already confirmed empty in earlier turns and directly located the target file discovered in turn 3. Production logs show the Selector adaptively adjusts selected turns (2 at round 3, 5 at round 6), consistent with offline analysis (\S\ref{sec:experiments}).

A sliding-window bug discovered during deployment (message loss after resume) inversely validates that Pull's context management is actively functioning---window miscalculation immediately causes LLM behavioral anomalies.

\subsection{Relation to Long-Context LLMs}

Why not simply use 1M-token context models? First, $O(N^2)$ cumulative retransmission cost means a larger window only postpones the cost collapse point rather than eliminating it (\S\ref{sec:intro}). Second, the signal-to-noise ratio is low: the entity-overlap median is only 20.7\%, so most of a larger window is still irrelevant history. Third, complementarity: Pull reduces prompt length, directly lowering KV cache prefill cost \citep{kwon2023pagedattention,zhang2023h2o,liu2023scissorhands}, and is orthogonal to recursive mechanisms like Transformer-XL \citep{dai2019transformerxl}.

\subsection{Pull as the Second Layer}

Table~\ref{tab:positioning} contrasts Pull with existing memory systems along the dimensions that matter for the second layer. The comparison is complementary rather than competitive: MemGPT/Letta, Mem0, and Zep target cross-session agent state and knowledge archival, whereas Pull targets intra-session working memory with deterministic, reversible materialization. ACE \citep{liao2026ace} likewise retains lossless storage with reversible expansion, but its maintenance layer is LLM-driven (LLM-generated abstractions, LLM elasticizer); the uniqueness claim here concerns the combination of deterministic zero-LLM maintenance with reversible turn-level materialization.

\begin{table}[t]
\centering
\small
\resizebox{\columnwidth}{!}{%
\begin{tabular}{@{}lllll@{}}
\toprule
\textbf{System} & \textbf{Unit} & \textbf{Routing decision} & \textbf{Reversible} & \textbf{Target layer} \\
\midrule
MemGPT/Letta & page & LLM (self-directed) & partial & agent state \\
Mem0 & fact & LLM extraction & lossy & cross-session \\
Zep & KG edge & temporal graph & lossy & cross-session \\
RAG & chunk & similarity & n/a & retrieval \\
\textbf{Pull} & turn & LLM over determ.\ metadata & \textbf{yes} & \textbf{intra-session} \\
\bottomrule
\end{tabular}}
\caption{Positioning of Pull relative to memory systems. Pull is the only design combining deterministic (zero-LLM) maintenance with fully reversible turn-level materialization for intra-session working memory.}
\label{tab:positioning}
\end{table}

Pull is not a replacement for RAG or long-term memory; it is a precondition. In a full production stack, Pull manages active working memory while evicted turns are archived to third-layer systems for cross-session recall. As formalized in \S\ref{sec:method} and Figure~\ref{fig:tiered}, this layered architecture mirrors classical OS memory hierarchies: registers (LLM context) $\to$ L1 cache (active zone) $\to$ L2 cache (frozen zone) $\to$ main memory (third-layer archive).

\section{Conclusion}

Pull is a session router that fills the missing second layer in LLM application architecture---session working memory management---via lazy materialization. On LoCoEval, Pull reduces Phase~2 tokens by 75.1\% (single-hop) and 72.0\% (multi-hop) with equivalent or improved quality. A controlled routing benchmark shows entity lifecycle tracking is empirically a prerequisite for distance-independent routing. BEAM~1M validates the paradigm on million-token conversations (+55.2\% F1 vs truncation). Pull's true value lies not in ``how many tokens are saved,'' but in defining a missing standard layer in LLM application architecture. This separation of concerns (deterministic local metadata + on-demand LLM materialization) offers a reusable architectural pattern.

\section*{Limitations}

\begin{enumerate}
    \item M0 summaries are extractive (no generative hallucination), but extreme summarization omissions may affect routing; token savings are model-dependent (75.1\% on GLM-5 batched vs 45.6\% on Qwen3.7plus per-task).
    \item Entity extraction and the controlled routing benchmark are code-domain-centric (SWE-bench traces); generalization to low-entity-density domains and other benchmarks is future work.
    \item Evaluation scope: the main experiment uses the same model (GLM-5) for generation and evaluation (independent Qwen3.7plus judge confirms directional consistency, $\Delta = +0.0148$, Pearson $r > 0.85$, but 22/64 samples show opposite judgments); BEAM uses 14/35 conversations with a simplified Purifier; LLMLingua2 comparison is multi-hop only.
    \item The introductory empirical findings derive from a single-user development log (287 sessions) and may not generalize to multi-user or non-coding settings. Production deployment uses no A/B control (all instances use Pull); token savings are modest in absolute terms (${\sim}$10K tokens/instance), with primary value in latency reduction and context-bloat prevention.
\end{enumerate}

\bibliography{refs}

\appendix

\section{Cross-Model Validation}

\begin{table}[H]
\centering
\small
\begin{tabular}{@{}lcc@{}}
\toprule
\textbf{Metric} & \textbf{GLM-5 (batch)} & \textbf{Qwen3.7plus (task)} \\
\midrule
Vanilla F1 & 0.6694 & 0.7201 \\
Pull F1 & 0.6676 & 0.7225 \\
$\Delta$ & $-$0.0018 & +0.0024 \\
Token savings & 75.1\% & 45.6\% \\
\bottomrule
\end{tabular}
\caption{Cross-model validation (64 conversations, single-hop).}
\label{tab:crossmodel}
\end{table}

\section{Independent Judge Validation}

\begin{table}[H]
\centering
\small
\begin{tabular}{@{}lccc@{}}
\toprule
\textbf{Judge} & \textbf{Vanilla} & \textbf{Pull} & $\Delta$ \\
\midrule
GLM-5 (same-src) & 0.6694 & 0.6676 & $-$0.0018 \\
Qwen3.7plus (indep.) & 0.7186 & 0.7334 & +0.0148 \\
\bottomrule
\end{tabular}
\caption{Independent judge validation (single-hop, 64 conversations). Pearson $r > 0.85$ for sample ordering; 42/64 samples show consistent $\Delta$ direction.}
\label{tab:judge}
\end{table}

\section{Selector Precision Analysis}

\begin{table}[H]
\centering
\small
\resizebox{\columnwidth}{!}{%
\begin{tabular}{@{}lcc@{}}
\toprule
\textbf{Model} & \textbf{Simplified} & \textbf{Full} \\
\midrule
GLM-5 & R=.27 P=.34 H=58\% & R=.38 P=.83 H=84\% \\
Qwen3.7plus & R=.78 P=.83 H=88\% & R=.42 P=.82 H=88\% \\
Kimi K2.5 & R=.67 P=.84 H=90\% & R=.40 P=.83 H=88\% \\
\bottomrule
\end{tabular}}
\caption{Selector key-turn recall (50 queries, macro average). Micro precision for GLM-5 + simplified: 381/466 = 81.8\%.}
\label{tab:selector}
\end{table}

\section{Pure M0 Ablation and Contribution Chain}

\begin{table}[H]
\centering
\small
\begin{tabular}{@{}lccc@{}}
\toprule
\textbf{Method} & \textbf{Single-hop} & \textbf{Multi-hop} \\
\midrule
Pure M0 & 0.5143 & 0.5672 \\
Full Pull & 0.6676 & 0.6500 \\
Vanilla & 0.6694 & 0.6326$^\dagger$ \\
\bottomrule
\end{tabular}
\caption{Pure M0 ablation. $^\dagger$32-sample subset.}
\label{tab:purem0}
\end{table}

\begin{table}[H]
\centering
\small
\begin{tabular}{@{}lccccc@{}}
\toprule
\textbf{Config} & \textbf{M0} & \textbf{Sel.} & \textbf{Casc.} & \textbf{F1} & $\Delta$ \\
\midrule
Pure M0 & \checkmark & $\times$ & $\times$ & 0.5143 & $-$0.153 \\
w/o M0 & $\times$ & \checkmark & \checkmark & 0.6103 & $-$0.059 \\
w/o Cascade & \checkmark & \checkmark & $\times$ & 0.6728 & +0.003 \\
\textbf{Full Pull} & \checkmark & \checkmark & \checkmark & \textbf{0.6676} & --- \\
\bottomrule
\end{tabular}
\caption{Contribution chain (single-hop, batched Answerer).}
\label{tab:contribution}
\end{table}

\section{Distance Decay and Layer Analysis}

\begin{table}[H]
\centering
\small
\begin{tabular}{@{}lcccc@{}}
\toprule
\textbf{Distance} & \textbf{M10} & \textbf{RAG+Ent} & \textbf{RAG-BM25} & \textbf{Last-K} \\
\midrule
$\leq$30 & 78.4\% & 60.2\% & 40.3\% & 75.6\% \\
31--80 & 98.0\% & 79.4\% & 57.2\% & 3.9\% \\
81--200 & 97.0\% & 83.6\% & 69.9\% & 9.0\% \\
$>$200 & 98.4\% & 70.6\% & 50.5\% & 0.8\% \\
\bottomrule
\end{tabular}
\caption{Recall@10 by distance (7{,}831 queries). RAG+Ent = RAG-BM25 augmented with structured entity fields (the +Entity row of Table~\ref{tab:routing_ablation}). M10 shows no distance decay; Last-K collapses. The slightly lower Recall at ${\leq}$30 steps (78.4\%) reflects higher query density in this band: short-range probes target fine-grained state changes where multiple candidate turns compete for the fixed budget, whereas distant queries have sparser signal and less ambiguity.}
\label{tab:distance}
\end{table}

\section{Latency Analysis}

\begin{table}[H]
\centering
\small
\begin{tabular}{@{}lccccc@{}}
\toprule
\textbf{Length} & \textbf{Full} & \textbf{Sel.} & \textbf{Ans.} & \textbf{E2E} & $\Delta$ \\
\midrule
10K & 1.19s & 1.01s & 0.77s & 1.78s & $-$0.59s \\
50K & 1.59s & 0.98s & 0.82s & 1.80s & $-$0.21s \\
100K & 2.09s & 1.07s & 0.93s & 2.00s & +0.09s \\
200K & 2.67s & 1.08s & 0.85s & \textbf{1.93s} & \textbf{+0.74s} \\
\bottomrule
\end{tabular}
\caption{End-to-end latency (GLM-5 API, streaming, 20 trials, p50). $\Delta = \text{Full} - \text{Pull E2E}$ (positive = Pull faster).}
\label{tab:latency}
\end{table}

\section{Archive and Frozen Zone Validation}

\begin{table}[H]
\centering
\small
\begin{tabular}{@{}lccc@{}}
\toprule
\textbf{Zone} & \textbf{Van. (181K)} & \textbf{Pull (35K)} & \textbf{No Arch. (29K)} \\
\midrule
Frozen & 0.000 & \textbf{0.473} & 0.030 \\
Active & 0.662 & \textbf{0.796} & 0.801 \\
Cross & 0.024 & \textbf{0.246} & 0.207 \\
\bottomrule
\end{tabular}
\caption{Archive validation (6 stitched conversations, ${\sim}$800 turns each). Token savings: 80.7\%.}
\label{tab:archive}
\end{table}

\section{BEAM 1M Retrieval Baselines}

\begin{table}[H]
\centering
\small
\begin{tabular}{@{}lccc@{}}
\toprule
\textbf{Method} & \textbf{Tokens} & \textbf{F1} & $\Delta$ vs Pull \\
\midrule
BM25 & 30{,}879 & 0.082 & $-$0.382 \\
Dense & 29{,}413 & 0.067 & $-$0.397 \\
Hybrid & 63{,}741 & 0.128 & $-$0.336 \\
BM25 (163K) & 163{,}077 & 0.137 & $-$0.327 \\
Dense (163K) & 163{,}077 & 0.109 & $-$0.355 \\
\textbf{Pull} & \textbf{64{,}517} & \textbf{0.464} & --- \\
\bottomrule
\end{tabular}
\caption{BEAM 1M retrieval baselines (263 questions). 163K variants use Last-K truncation at the same token budget.}
\label{tab:beam_retrieval}
\end{table}

\section{Failed Alternatives}

\textbf{LLM-driven entity extraction:} per-turn latency from 3.2 ms to ${\sim}$2 s, with non-determinism. \textbf{Generative M0 summaries:} hallucination risk; 0.08 F1 lower than extractive M0 in 16-sample pilot. \textbf{3-hop cascade:} 40\% token increase for only +0.002 F1. \textbf{Selector generating answers directly:} $-$0.15 F1.

\section{Purifier Detail}
\label{sec:purifier-detail}

\paragraph{Entity template families.} The extractor compiles 16 regex patterns that collapse into 13 type families (the three date variants and two capitalized-name variants each count as one family). Table~\ref{tab:templates} lists each family with a representative pattern.

\begin{table}[h]
\centering
\footnotesize
\resizebox{\columnwidth}{!}{%
\begin{tabular}{@{}ll@{}}
\toprule
\textbf{Family} & \textbf{Representative pattern} \\
\midrule
function/class def & \texttt{\textbackslash b(def|class)\textbackslash s+(\textbackslash w+)} \\
snake\_case ident & \texttt{[a-z]+(\_[a-z]+)+} \\
camelCase ident & \texttt{[a-z]+([A-Z][a-z]*)\{2,\}} \\
backtick ident & \texttt{`(\textbackslash w+)`} \\
URL & \texttt{https?://\textbackslash S+} \\
@mention & \texttt{@\textbackslash w+} \\
\#hashtag & \texttt{\#\textbackslash w+} \\
number/currency/pct & \texttt{\textbackslash d+}, currency prefix, \texttt{\textbackslash d+\%} \\
date (EN/ISO/ZH) & month-name / ISO / CJK \\
time & \texttt{\textbackslash d\{1,2\}:\textbackslash d\{2\}} \\
capitalized name & filtered proper nouns \\
\bottomrule
\end{tabular}}
\caption{The 13 entity template families (16 underlying patterns; 11 rows shown after merging date and capitalized-name variants). Identifiers carry minimum-length thresholds (camelCase $\geq$6, snake\_case $\geq$8 chars); all nuggets are lower-cased for case-insensitive matching.}
\label{tab:templates}
\end{table}

\paragraph{Fidelity (information-density) scorer.} The fidelity score $d\in[0,2]$ is a hand-weighted sum of nine signals: five boolean pragmatic detectors (entity $+0.5$, intent $+0.4$, constraint $+0.5$, question $+0.3$, alignment $+0.2$, each an EN+ZH regex), content-keyword density ($+0.05$ per keyword), normalized character entropy ($+0.06$), rare-word bonus ($+0.04$ each, capped at $0.2$), and a structured-text bonus ($+0.2$). Two multiplicative penalties apply: $\times0.5$ for texts shorter than 10 chars and $\times0.3$ for templated agent acknowledgements; the result is capped at $2.0$. A turn is KEEP/COMPRESS/FILTER-classified by thresholding $d$ (aggressive mode: $d<0.35\to$FILTER, $d<0.65\to$COMPRESS), with hard rules forcing KEEP when any pragmatic signal fires.

\section{Selector Prompt and M10 Format}
\label{sec:selector-prompt}

\paragraph{Selector system prompt.} The Selector is instructed to select turns, never to answer. The full system prompt (path example anonymized):

\begin{mdframed}[style=codestyle]
\scriptsize
\begin{verbatim}
YOU ARE A TURN SELECTOR, not an answerer.

You are reviewing a compressed index
of the conversation history. Each line
is one turn -- you MUST select which
turns to EXPAND so a downstream
answerer can read their content.

Index fields:
  path       -- file path (src/module.py)
  entities   -- identifiers in this turn
  [def:X]    -- X first defined here
  [mod:X]    -- X modified here
  [type_tag] -- issue/code/test/err/think
  d=H/M/L    -- information density
  keep:XX%   -- estimated retention
  grade:     -- HIGH / MEDIUM / LOW
  [pivot:..] -- topic shift marker
  [branch:X] -- forked branch

RULES:
1. MUST select turns. ALWAYS output nums.
2. Never say "NONE" unless info absent.
3. Do NOT answer. Do NOT summarize.
4. Select only the 3-5 MOST RELEVANT.

Output:
Reason: <one sentence: which and why>
TURNS: <comma-separated numbers>
\end{verbatim}
\end{mdframed}

\noindent The vocabulary also declares a \texttt{[dep:X]} (deprecated) marker; the current extractor emits \texttt{[def]} and \texttt{[mod]} in M10 (deprecation is tracked internally for eviction but not rendered to the Selector).

\paragraph{M10 input format.} M10 is structured plain text, one line per turn (not JSON). Two representative lines:

\begin{mdframed}[style=codestyle]
\scriptsize
\begin{verbatim}
12: [code] src/order_repo.py OrderRepo
    save_order [def:OrderRepo] d=H
    saved:42% keep:78% grade:HIGH
    Defined OrderRepo with save_order
27: [think] OrderRepo save_order
    [mod:save_order] d=M saved:30%
    keep:55% grade:MEDIUM [pivot:
    save_order drops null fields]
    Realized save_order drops nulls
\end{verbatim}
\end{mdframed}

\noindent Field order: \texttt{turn\_id}, type tag, path, entities ($\leq$5), lifecycle markers, density, savings, retention/grade, pivot/branch markers, and finally the ${\sim}$100-char extractive summary (placed last, ``identity-first'').

\paragraph{Output parsing.} The LLM emits two lines (\texttt{Reason:} + \texttt{TURNS:}). The parser is fault-tolerant: it tries, in order, \texttt{TURNS: NONE}$\to[]$, \texttt{TURNS: 2,5,7}, a bare comma-separated list, and a JSON fallback; all IDs are range-checked ($0\leq t<N$), de-duplicated, and sorted. On total failure it falls back to the last three turns.

\paragraph{Cascade expansion (rule-based).} Cascade is deterministic post-processing applied after the LLM output, completing causal antecedents by entity co-reference:

\begin{mdframed}[style=codestyle]
\scriptsize
\begin{verbatim}
def cascade(S0, max_extra=5):
    first = {e: earliest turn defining e}
    C = { first[e]
          for t in S0
          for e in entities(t)
          if first[e] not in S0 }
    return sorted(C)[:max_extra]
\end{verbatim}
\end{mdframed}

\noindent Expansion is one hop (no transitive closure) and bounded by a fixed budget of five extra turns ($|S|\leq|S_0|+5$). It is ablated by replacing it with the empty function (the \textit{w/o Cascade} variant).

\section{Failure Case Studies}
\label{sec:casestudy}

To characterize where Pull fails, we examine three single-hop conversations (repository names anonymized as Repo-A/B/C) from the main evaluation set, each isolating one mechanism. Figures are F1 under each ablation; pull counts are the number of materialized turns.

\paragraph{Case 1: Cascade over-expansion displaces seed turns (Repo-A).}
F1: vanilla 0.672, full 0.390, w/o cascade 0.668, w/o M0 0.648. Pull materializes 19 turns---fewer than w/o cascade's 21---yet F1 collapses; removing cascade alone restores vanilla-level quality.
\begin{mdframed}[style=quotestyle]\scriptsize\raggedright\hyphenpenalty=10000\exhyphenpenalty=10000
T31 (user): ``Add \texttt{fetch\_permission\_principals} to \texttt{InMemoryPermission} in \texttt{memory\_backend.py}.'' \textit{(cascade anchor)}\\
T21 (user): ``How do I run the test suite locally with tox?'' \textit{(causally linked, task-irrelevant)}
\end{mdframed}
\begin{mdframed}[style=codestyle]
\scriptsize
\begin{verbatim}
T31 [think] fetch_permission_principals,
    InMemoryPermission
    [def:fetch_permission_principals] d=H
    "Proposes fetch_permission_..."
T21 [think] tox, dev-requirements d=L
    "Asks how to run the test suite"
\end{verbatim}
\end{mdframed}
{\raggedright\noindent\textit{Root cause.} Cascade fired on \texttt{fetch\_permission\_principals} and expanded backward into a contiguous block (T20--33) of causally linked but largely irrelevant turns. Under the fixed pull budget this displaced the dispersed seed turns retained by the w/o-cascade router; the answerer then mis-attributes the unrelated \texttt{deserialize} function to \texttt{memory\_backend.py} (vanilla and w/o-cascade place it correctly in \texttt{schema.py}). The failure is isolated to cascade: removing it restores F1 (0.390$\to$0.668).\par}

\paragraph{Case 2: ${\sim}$100-char M0 summaries collapse distinct drafts (Repo-B).}
F1: vanilla 0.635, full 0.476, w/o cascade 0.632, w/o M0 0.705. Removing the M0 summary (full text) yields the \textit{best} F1, surpassing vanilla.
\begin{mdframed}[style=quotestyle]\scriptsize\raggedright\hyphenpenalty=10000\exhyphenpenalty=10000
T49 (early wrong draft): ``For \texttt{UploadDirManager.add}, check \texttt{is\_cloud\_uri} and cache the auto-name.''\\
T67 (corrected draft): ``Use a \texttt{\_names\_taken} set; the cache is \texttt{\_path\_to\_name}; check \texttt{is\_uri}.''
\end{mdframed}
\noindent\textit{Root cause.} The \texttt{add} method is revised across many turns (T49$\to$T67); the correct final design differs from early drafts only in fine-grained identifier choices. A ${\sim}$100-char M0 summary collapses both drafts to roughly ``discusses \texttt{add} URI check + unique name,'' so the router cannot distinguish the authoritative turn from superseded drafts and spends budget on topically salient but irrelevant turns (an SSH-tunnel discussion). Giving the router full text (w/o M0) lets it select the refined turns, lifting F1 to 0.705---the best of all conditions. The failure is attributable to M0 lossiness, not cascade (w/o cascade stays at 0.632).

\paragraph{Case 3: Regex extractor never recalls the task-critical cluster (Repo-C).}
F1: vanilla 0.720, full 0.575, w/o cascade 0.633, w/o M0 0.681. The pull sets are nearly identical across ablations (27/25/24 turns); \textit{none} contains the T75--104 cluster defining both \texttt{connect\_to\_region} tasks plus the multi-hop task.
\begin{mdframed}[style=quotestyle]\scriptsize\raggedright\hyphenpenalty=10000\exhyphenpenalty=10000
T75 (never pulled): ``Implementing \texttt{connect\_to\_region} in \texttt{service\_x}, delegating to a region factory.''\\
T93 (never pulled): ``The \texttt{connect\_to\_region} in \texttt{service\_y}---is \texttt{KinesisConnection} right?''
\end{mdframed}
\noindent\textit{Root cause.} The regex extractor seeded on surface entities it could match (\texttt{compute\_md5}, SDK-version diffs) but never recalled the \texttt{connect\_to\_region} definition cluster (T75--104), which underlies two of four tasks plus the multi-hop task. With those turns absent, the answerer conflates the two services and swaps their file paths---a direct swap error. Because the seed was never surfaced, toggling cascade (0.633) or M0 (0.681) cannot recover vanilla (0.720); all ablations plateau at nearly the same pull count, isolating the failure to the entity-recall stage.

\paragraph{Takeaway.} The three failures map cleanly onto the three mechanisms: cascade can over-expand into contiguous irrelevant blocks (bounded by the fixed budget but not eliminated), M0 summarization can collapse fine-grained distinctions between revised drafts, and regex entity recall has a coverage ceiling in low-signal regions. Each is consistent with the corresponding Limitation and points to a concrete future direction (learned cascade scoring, draft-aware summaries, hybrid neural entity recall).

\end{document}